\documentclass[sigconf,nonacm]{acmart}
\AtBeginDocument{%
  }

\acmConference[DAC'26 under submission]{Note}{July 26-29, 2026}{Long Beach  CA}

\usepackage{multirow}
\usepackage{subcaption}
\usepackage{algorithm}
\usepackage{newfloat}
\usepackage{listings}
\usepackage{booktabs} %
\usepackage{threeparttable}
\usepackage{algpseudocode}
\usepackage{wasysym}
\usepackage{tikz}
\usepackage{mathtools}

\newcommand{\blackcircle}[1]{%
  \tikz[baseline=(char.base)]{
    \node[shape=circle,fill=black,inner sep=1pt] (char)
      {\color{white}\bfseries #1};
  }%
}

\begin{document}

\title{GSR-GNN: Training Acceleration and Memory-Saving Framework of Deep GNNs on Circuit Graph}

\author{Yuebo Luo}
\email{luo00466@umn.edu}
\orcid{1234-5678-9012}
\affiliation{%
  \institution{University of Minnesota, Twin Cities}
  \city{Minneapolis}
  \state{Minnesota}
  \country{USA}
}

\author{Shiyang Li}
\email{li004074@umn.edu}
\affiliation{%
  \institution{University of Minnesota, Twin Cities}
  \city{Minneapolis}
  \state{Minnesota}
  \country{USA}
}

\author{Yifei Feng}
\email{yf3005@nyu.edu}
\affiliation{%
  \institution{New York University}
  \city{Brooklyn}
  \state{New York}
  \country{USA}
}

\author{Vishal Kancharla}
\email{kanch042@umn.edu}
\affiliation{%
  \institution{University of Minnesota, Twin Cities}
  \city{Minneapolis}
  \state{Minnesota}
  \country{USA}
}

\author{Shaoyi Huang}
\email{shuang59@stevens.edu}
\affiliation{%
  \institution{Stevens Institute of Technology}
  \city{Hoboken}
  \state{New Jersy}
  \country{USA}}

\author{Caiwen Ding}
\email{dingc@umn.edu}
\affiliation{%
  \institution{University of Minnesota, Twin Cities}
  \city{Minneapolis}
  \state{Minnesota}
  \country{USA}
}




\begin{abstract}
Graph Neural Networks (GNNs) show strong promise for circuit analysis, but scaling to modern large-scale circuit graphs is limited by GPU memory and training cost, especially for deep models. We revisit deep GNNs for circuit graphs and show that, when trainable, they significantly outperform shallow architectures, motivating an efficient, domain-specific training framework. We propose Grouped-Sparse-Reversible GNN (GSR-GNN), which enables training GNNs with up to hundreds of layers while reducing both compute and memory overhead. GSR-GNN integrates reversible residual modules with a group-wise sparse nonlinear operator that compresses node embeddings without sacrificing task-relevant information, and employs an optimized execution pipeline to eliminate fragmented activation storage and reduce data movement. On sampled circuit graphs, GSR-GNN achieves up to 87.2\% peak memory reduction and over 30$\times$ training speedup with negligible degradation in correlation-based quality metrics, making deep GNNs practical for large-scale EDA workloads.
\end{abstract}

\keywords{GNN, EDA, Circuit Design, Reversible Computing }


\maketitle

\section{Introduction}

Graph Neural Networks (GNNs) have shown strong and consistent performance in circuit graph analysis~\cite{Kaufmann+2022+285+291, 8105885}. For example, verifying a Booth multiplier can take over 100 hours using the commercial OneSpin equivalence checking tool~\cite{7459464}, while GNNs can finish the same task in less than one minute. 

 \begin{figure}[t]
     \centering
     \includegraphics[width=\linewidth]{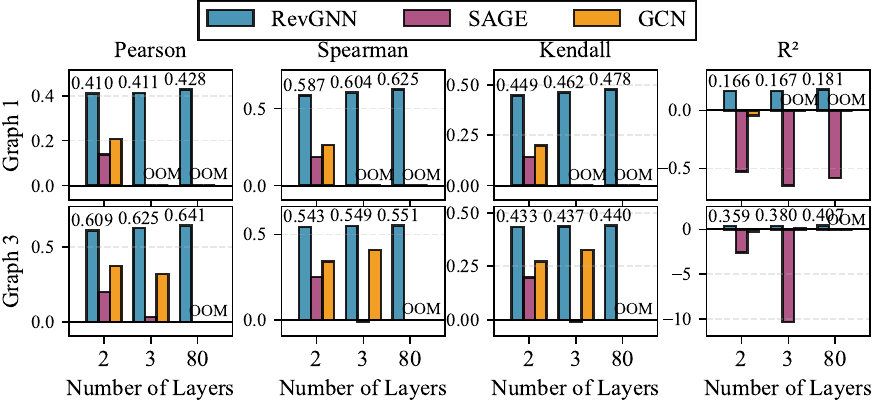}
     \caption{Performance Comparison: RevGNN vs. SAGE vs. GCN on Circuit Graphs 1 and 3 (See Table \ref{tab:compact_graph_summary}), where RevGNN outperforms SAGE and GCN and keeps gaining return when increasing depth. Negative scores in Spearman, Kendall, and $R^2$ mean failure to find correlations.}
     \label{fig:revgnn-perf-comp}
\end{figure}
With ongoing technological advances, circuit graphs, graph data extracted from circuit designs, are becoming significantly larger and more complex~\cite{microsoftgraph,hu2020open,chowdhury2021openabcdlargescaledatasetmachine,10158384,jiang2024circuitnet} with greatly varied connection patterns.  This growth has pushed traditional shallow GNNs—typically 3–4 layers---to their performance limits~\cite{10.1145/3489517.3530597,zhang2021evaluatingdeepgraphneural}.

Prior research has paved the way for training extremely deep GNNs, reaching hundreds or even thousands of layers. RevGNN~\cite{revgnn}, for instance, introduces a group-wise reversible residual structure. It significantly reduces the memory complexity of GNNs from $O(LND)$ (where $L$ is depth, $N$ is nodes, $D$ is features) to $O(ND)$, effectively making it independent of depth. 
Our observations in Figure   \ref{fig:revgnn-perf-comp} show that training deep RevGNN leads to significant performance improvements in downstream circuit design tasks compared to conventionally shallow GNN architectures, where RevGNN's Pearson can be pushed to 0.4277 at 80 layers from 0.4096 at two layers, whereas  GCN~\cite{gcn} and GraphSAGE~\cite{graphsage} stay below 0.3731 at best or even zero scores when there are two layers. 
 \begin{figure}[t]
    \centering
    \includegraphics[width=1\linewidth]{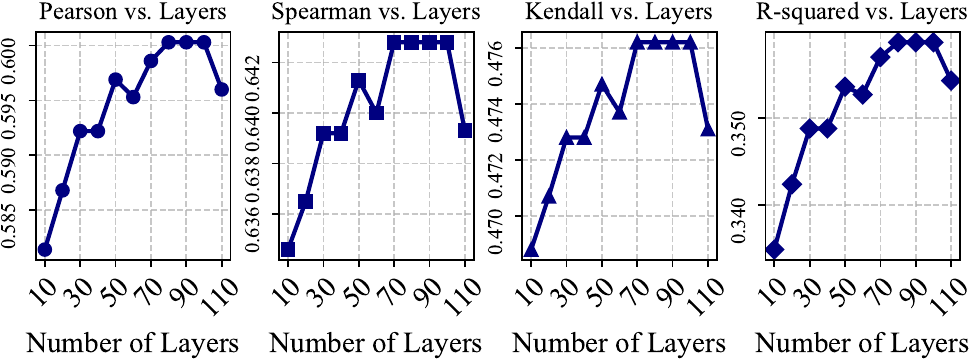}
    \caption{Training results of baseline RevGNN on example graph 356-R in Table \ref{tab:compact_graph_summary}, where RevGNN pleateaus around 70 layers, where SAGE and GCN are either too low to stay in the range or crashed because of out-of-memory (OOM) error.}
    \vspace{-10pt}
    \label{fig:example-graph-metrics}
\end{figure}

\begin{figure}[t]
    \centering
    \includegraphics[width=0.95\linewidth]{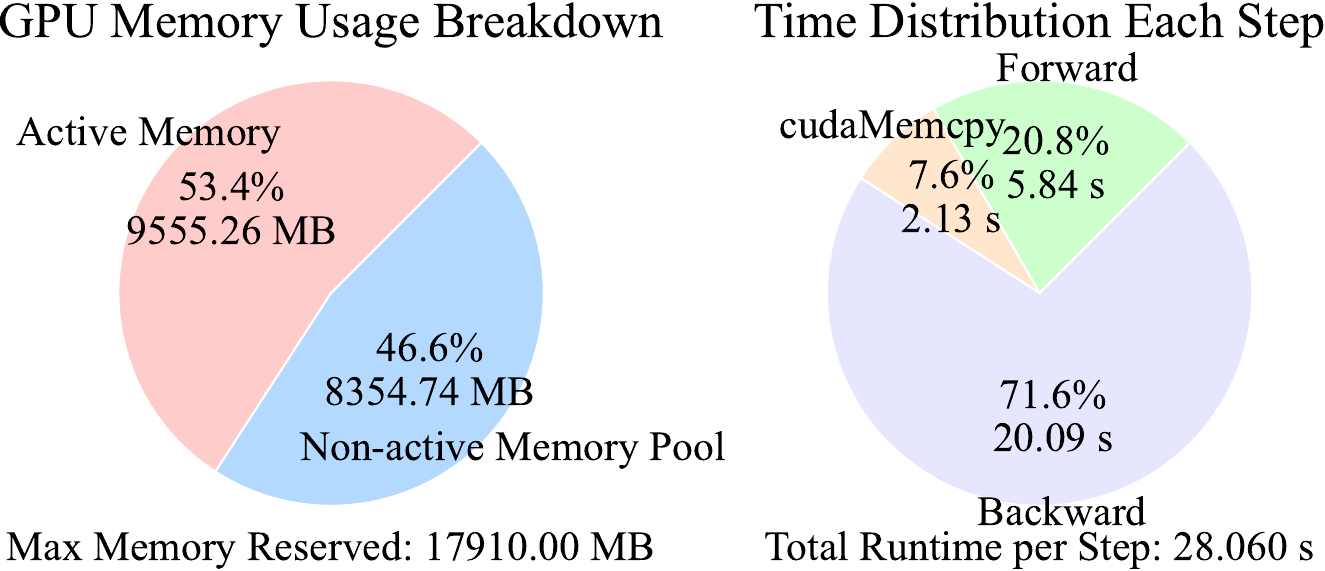}
    \caption{GPU memory usage and epoch runtime summary of 100-layer RevGNN with 384 hidden channels on Graph 356-RISCY-a1-c5 in Table \ref{tab:compact_graph_summary}. }
    \label{fig:revgnn-memory-time}
       \vspace{-10pt}
\end{figure}

\begin{figure}[tt]
    \centering
    \includegraphics[width=0.99\linewidth]{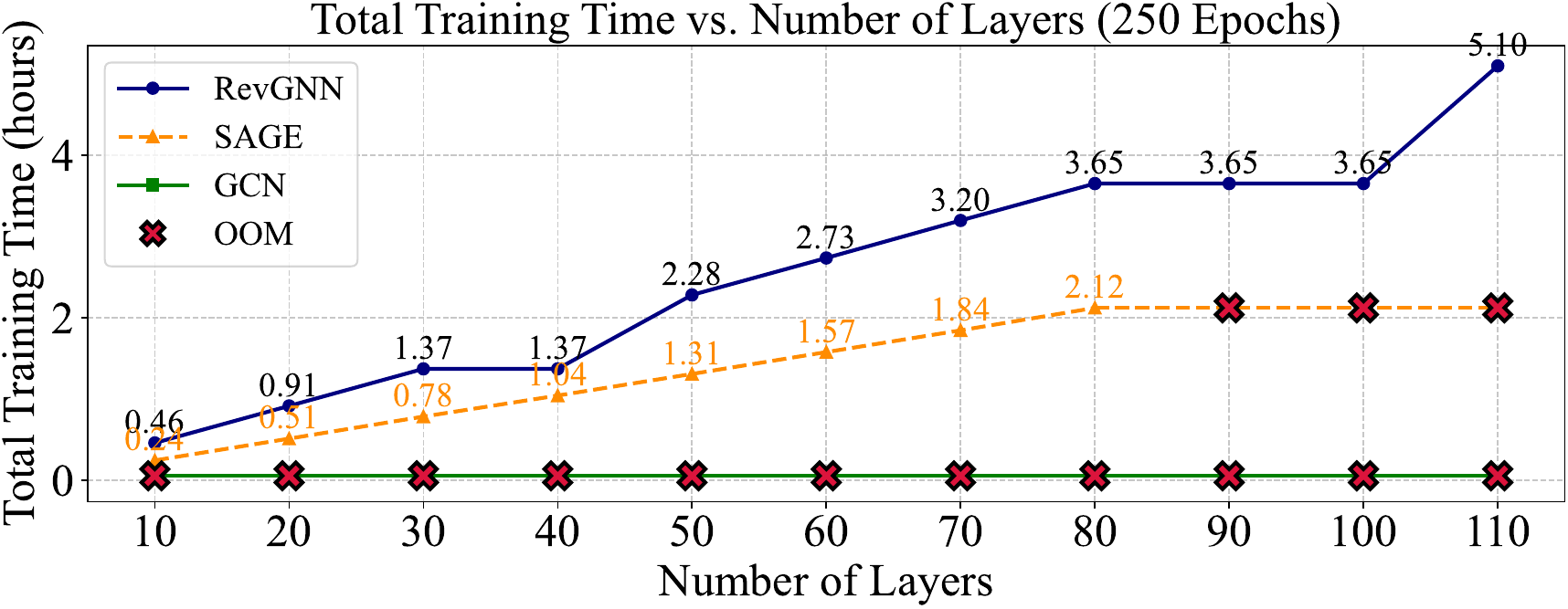}
    \caption{Example training time of baseline 100-layer RevGNN on graph 7 in Table \ref{tab:compact_graph_summary}. Note that the red 'X' refers to an out-of-memory (OOM) error.}
    \label{fig:example-graph-training-time}
    \vspace{-20pt}
\end{figure}

Nonetheless, when approaching its theoretical potential in circuit design domains, deep RevGNN hits a scalability bottleneck rooted in its chunk-based, create-then-destroy memory management strategy. Our profiling of RevGNN on the CircuitNet dataset~\cite{jiang2024circuitnet} reveals that this memory management approach creates a cascade of practical limitations: (i) excessive GPU memory reservations (e.g., 17910 MB reserved with only 9555.26 MB active, as shown in Figure \ref{fig:revgnn-memory-time}), prohibitively long training times (5.10 hours for 250 epochs on a single circuit graph 356-R selected from Table \ref{tab:compact_graph_summary}, shown in Figure \ref{fig:example-graph-training-time}), and eventual performance degradation in all four correlation scores: Pearson, Spearman, Kendall, and $R^2$ beyond optimal depth at around 70 layers (please see Figure \ref{fig:example-graph-metrics}). Collectively, these issues make RevGNN practically difficult for real-world circuit congestion prediction tasks. It requires up to 46.4 GPU hours and up to 30+ GB of GPU memory to train a 100-layer, 384-hidden-channel model on just 10 sampled graphs, while the entire dataset has 10,370 different circuit graphs in total.

To overcome these challenges, we streamline RevGNN's memory management approach, introduce efficient embedding compression with the relative sparse message passing carried by its GNN blocks, and eventually, propose our contributions to the solution as follows:

\begin{itemize}


\item To boost training efficiency, we introduce Grouped-Sparse-Reversible GNN (GSR-GNN), a high-speed and memory-efficient domain-specific deep GNN framework. GSR-GNN leverages a reversible residual structure to enable deep GNN training. We further accelerate training by incorporating embedding compression during both forward and backward passes, without negatively impacting model accuracy.

\item By optimizing the memory management workflow for deep GNN training, we've achieved substantial improvements in memory efficiency, maintaining over $90\%$ memory utilization and occupying much lower overall memory usage. Our framework also strikes an excellent balance between memory usage and training speed.

\item We profile our framework against the domain-specific circuit graph dataset CircuitNet. The evaluation results highlight our framework's breakneck training speed and memory-saving capabilities in deep GNN training for circuit graph datasets compared to baselines.

\end{itemize}

Experiment results demonstrate that GSR-GNN achieves up to above $35\times$ speedup, GPU memory saving of $87.22\%$, and $98.6\%$ of reserved memory is active over RevGNN on sampled circuit graph datasets when performing congestion prediction tasks~\cite{congestionpredictionnet2019,highdefcongestion,paintingonplacement,congestionaware}.





 \begin{figure}[t]
     \centering
     \includegraphics[width=0.95\linewidth]{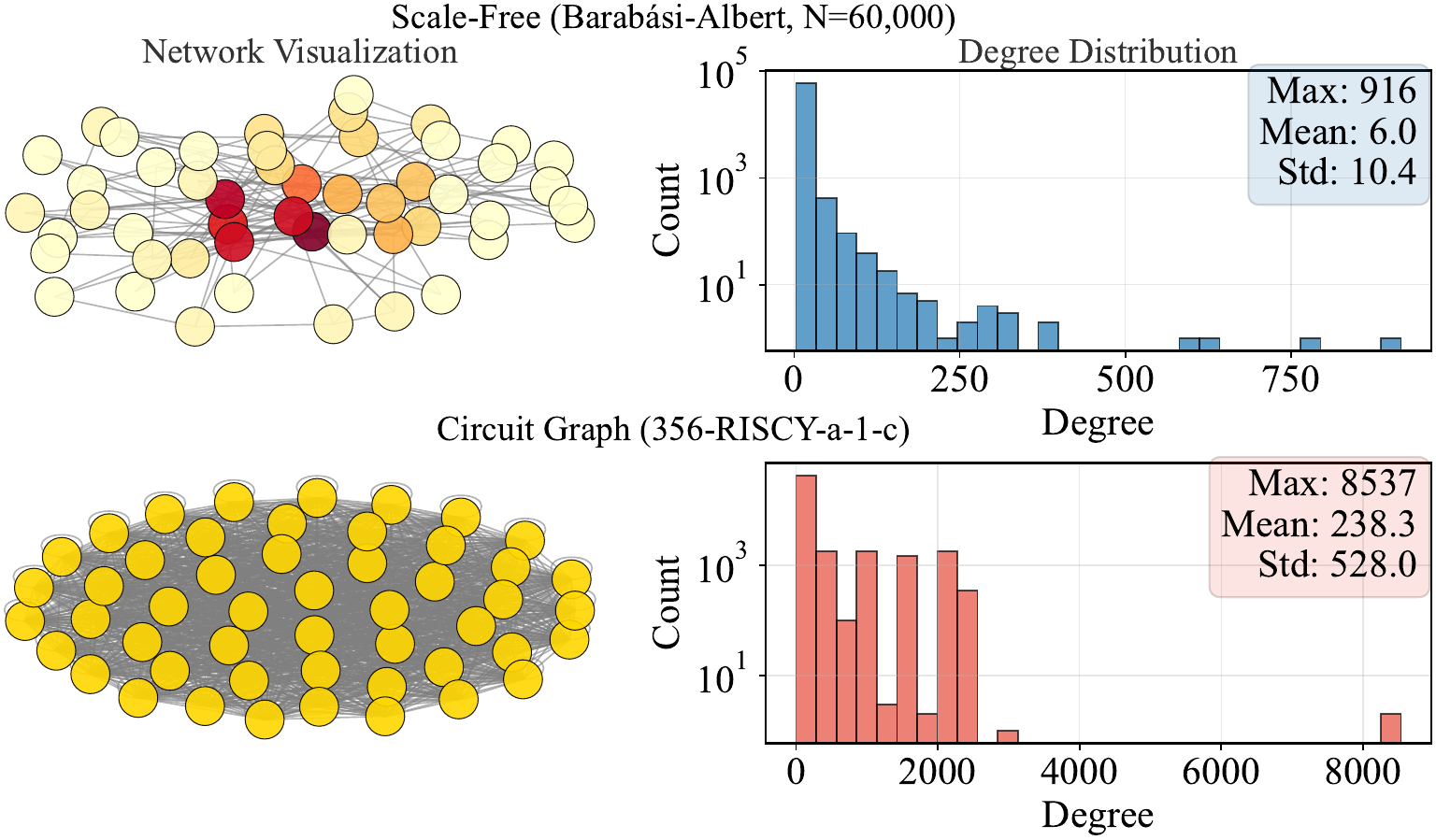}
     \caption{Comparison of Conventional graph dataset (upper) and circuit graph 4 (lower), Table \ref{tab:compact_graph_summary} in 50-node graph component pattern and node degree distribution. The circuit graph shows an apparently different connecting pattern and node degree distribution from the conventional graph.}
     \label{fig:circuit-graph-node-deg}
     \vspace{-0.45cm}
\end{figure}

\section{Background}


\textbf{Graph Learning in Circuit Design:} Circuit design workflows generate complex circuit representations ideal for graph-based learning~\cite{9598835,Dreamplace}. In contrast to a universal graph dataset such as the Barabási–Albert~\cite{Albert_2002}, shown in Figure \ref{fig:circuit-graph-node-deg}, circuit graphs exhibit unique characteristics: massive nodes, more bi-extreme towards either a few (as few as one) or many neighbours (more than 8,000) per node. Almost even and regular connectivity patterns in the microscope, along with diverse feature distributions, challenge conventional GNN implementations in this domain. 

Recent efforts have produced comprehensive circuit datasets~\cite{jiang2024circuitnet, 10158384} containing thousands of designs for tasks such as congestion prediction~\cite{congestionpredictionnet2019, generalizable}, Design Rules Check (DRC), timing estimation, and placement optimization, enabling the realistic evaluation of GNN efficiency improvements in terms of computation speedup and end-to-end performance.

\noindent \textbf{Reversible GNNs:} On top of the residual connection architecture to further the depth of GCN ~\cite{dinh2015nicenonlinearindependentcomponents,li2019deepgcns,li2020deepergcn},
RevGNN~\cite{revgnn} addresses a fundamental memory bottleneck in training deep GNNs. Traditional deep GNNs suffer from prohibitive memory overhead that scales linearly with the number of layers, as intermediate activations must be stored for backpropagation. This limits the practical depth of GNNs. RevGNN introduces a reversible architecture~\cite{gomez2017reversible,revgnn}, where the input to each layer can be reconstructed from its output during the backward pass. By eliminating the need to store intermediate activations, RevGNN makes memory complexity independent of network depth, significantly reducing the memory overhead during training. This breakthrough enables training GNNs with hundreds of layers on a single GPU.



\begin{figure*}[ht]
     \centering
     \includegraphics[width=1.0\linewidth]{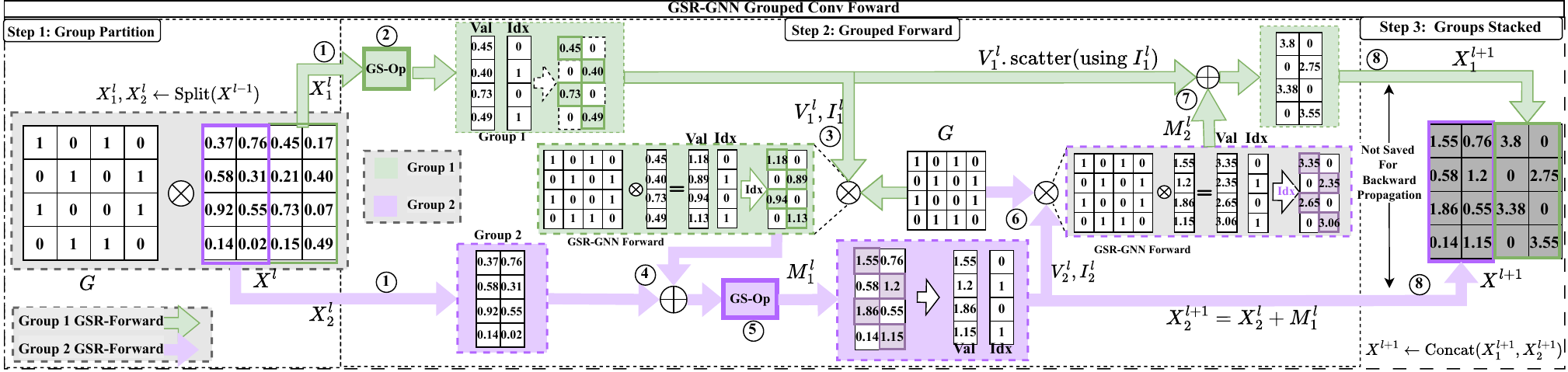}
     \caption{GSR-GNN in-layer forward workflow.}
     \label{fig:GSR-GNN-Fwd}
\end{figure*}

\section{Problem Statement and Challenges}
In this section, we first introduce the necessary terminology and notations used in the paper. 
Then, we present our observations that challenge current RevGNNs. 
Given graph $G = (V, E)$ with adjacency matrix $A \in \mathbb{R}^{N \times N}$, node embeddings at layer $l$ are denoted as $X^l \in \mathbb{R}^{N \times D}$, where $N = |V|$ and $D$ is the feature dimension. Network depth is $L$ with layer index $l \in \{1, 2, \ldots, L\}$, producing final predictions $\hat{y} = f_{\theta}(X^L) \in \mathbb{R}^N$.
During backpropagation, gradients flow as $\frac{\partial \mathcal{L}}{\partial \hat{y}} \rightarrow \frac{\partial \mathcal{L}}{\partial X^L} \rightarrow \cdots \rightarrow \frac{\partial \mathcal{L}}{\partial X^l}$. The $\text{Split}$ operation partitions tensors along the feature dimension, followed by Conventional GNN operations that process full-dimensional node embeddings and gradients:
{\footnotesize
\begin{align}
X_1^l, X_2^l &= \text{Split}(X^l) \quad \text{where } X_1^l, X_2^l \in \mathbb{R}^{N \times D/2} \\
X^{l+1} &= \text{GNN-Block-Forward}(G, X^l) \\
\frac{\partial \mathcal{L}}{\partial X^l} &= \text{GNN-Block-Backward}(G^T, \frac{\partial \mathcal{L}}{\partial X^{l+1}})
\end{align}
}
where forward pass aggregates neighbor information over $G$, and backward pass propagates gradients through $G^T$.

\subsection{Challenges for Reversible GNNs} 

Despite theoretical advantages, reversible GNN faces significant practical limitations that hinder its adoption in real-world applications. We identify three critical challenges that motivate the need for more efficient alternatives.

\noindent \textbf{Prohibitive Time-for-Space Tradeoff:} 
RevGNN adopts dynamic memory management by partitioning input features $\mathbf{X} \in \mathbb{R}^{N \times D}$ into $C$ groups, where $N = |\mathcal{V}|$ is the number of vertices and $D$ is the hidden dimension. This approach employs a \textit{create-then-destroy} cycle, in terms of detach, copy, and then delete at each layer $t \in [1, L]$ that incurs significant overhead. During forward propagation, activations are immediately deallocated after use:
{\footnotesize
\renewcommand{\arraystretch}{0.8}
\begin{align}
\mathcal{M}_{\text{input}}^{(t)} &\leftarrow \emptyset \quad \text{(deallocate via storage.resize(0))}, \quad \text{Peak}(\mathcal{M}) = \mathcal{O}\left(\frac{N \cdot D}{C}\right)
\end{align}
}
During backward propagation, the destroyed inputs must be reconstructed via memory reallocation and invertible computation:
{\footnotesize
\renewcommand{\arraystretch}{0.8}
\begin{align}
\mathcal{M}_{\text{input}}^{(t)} &\leftarrow \text{resize}\left(\prod_i d_i\right) = \text{malloc}(N \cdot D \cdot \text{sizeof(float32)}), \quad \mathbf{X}_{\text{input}}^{(t)} \leftarrow f^{-1}(\mathbf{Y}^{(t)})
\end{align}
}
The frequent allocation-deallocation cycle introduces three sources of overhead: (1) system calls for memory management, (2) memory fragmentation reducing cache efficiency, and (3) the invertible computation itself. The grouped reversible architecture computes:
{\footnotesize
\renewcommand{\arraystretch}{0.8}
\begin{align}
\text{Forward:} \quad &\mathbf{Y}_0' = \sum_{j=2}^{C} \mathbf{X}_j, \quad \mathbf{Y}_i' = f_{\theta_i}(\mathbf{Y}_{i-1}', \mathbf{A}) + \mathbf{X}_i, \quad i \in \{1, \ldots, C\} \\
\text{Inverse:} \quad &\mathbf{X}_i = \mathbf{Y}_i' - f_{\theta_i}(\mathbf{Y}_{i-1}', \mathbf{A}), \quad \mathbf{Y}_{i-1}' = \begin{cases} \mathbf{Y}_{i-1}' & i > 1 \\ \sum_{j=1}^{C} \mathbf{X}_j & i = 1 \end{cases}
\end{align}
}
where each partition $\mathbf{X}_i \in \mathbb{R}^{N \times D/C}$ and $f_{\theta_i}$ is a GNN block. The inversion requires reverse iteration through all $C$ groups. Aggregating over $L$ layers, without considering the edge embedding, the total computational complexity is:
{\footnotesize
\renewcommand{\arraystretch}{0.8}
\begin{align}
\mathcal{C}_{\text{total}} = \mathcal{O}\left(L \cdot \left(N \cdot \frac{D^2}{C}\right)\right) + \Theta(L)
\end{align}
}
where $\Theta(L)$ represents the cumulative allocation overhead. The timing analysis reveals a severe imbalance, with the total epoch time decomposed as:
{\footnotesize
\renewcommand{\arraystretch}{0.8}
\begin{align}
T_{\text{epoch}} &= T_{\text{forward}} + T_{\text{copy}} + T_{\text{backward}} 
\end{align}
}

\noindent\textbf{Empirical Analysis:} On a representative EDA benchmark, forward propagation requires $T_{\text{forward}} = 5.84s$ ($20.9\%$), memory copy (cudaMemcpy) takes $T_{\text{copy}} = 2.13s$ ($7.6\%$), while invertible backpropagation with repeated reallocation dominates at $T_{\text{backward}} = 20.02s$ ($71.6\%$), yielding $T_{\text{epoch}} = 27.99s$. The backward-to-forward ratio is $3.43$, and the numerical difference between $T_{\text{forward}}$ and $T_{\text{backward}}$ is $\Delta T \approx 14.18s$ per epoch. The backward-to-copy ratio of $9.40$ further highlights the computational burden from the create-destroy mechanism. This disproportionate cost can increase total training time by $10^1$ to $10^2\times$ for deep GNN models ($L > 50$), making RevGNN impractical for time-sensitive EDA applications (e.g., congestion prediction) where rapid design iteration is critical.

\noindent \textbf{Diminishing Gain from Extreme Depth:} The performance benefits of very deep RevGNNs plateau rapidly, particularly on domain-specific graphs. Figure~\ref {fig:example-graph-metrics} shows that correlation metrics on circuit graphs plateau around 70 layers and deteriorate beyond 100 layers. This indicates that moderate network depth achieves near-optimal performance, questioning the practical value of extreme depth capabilities while compounding the memory and computational efficiency challenges described above.


\begin{figure*}[ht]
     \centering
     \includegraphics[width=1.0\linewidth]{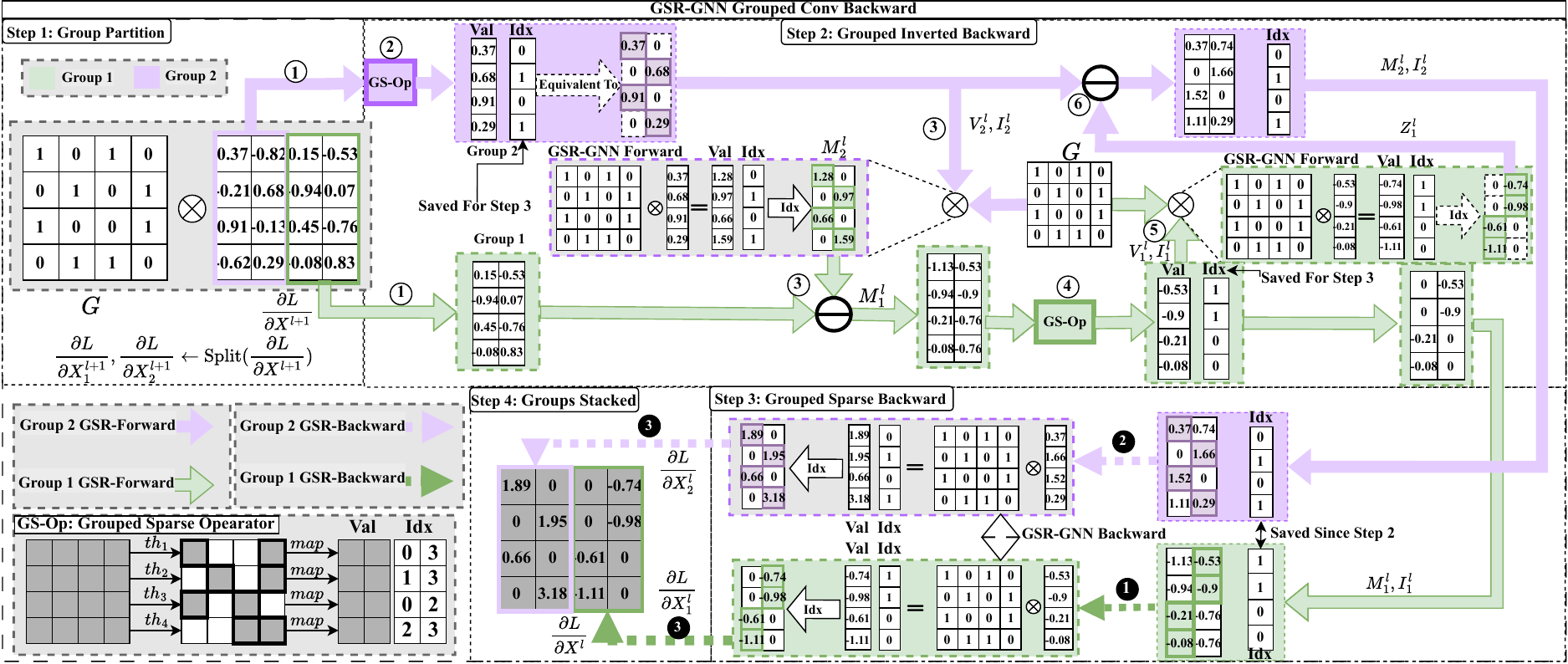}
     \caption{GSR-GNN in-layer backward workflow.}
     \label{fig:GSR-GNN-Arch-Bckwd}
\end{figure*}

\section{Architecture of GSR-GNN}
In this section, we present our general model architecture design, including our custom operators utilized in GSR-GNN, optimized memory management workflow, and our algorithm design in the forward and backward phases. 

\subsection{Operators and Blocks used in GSR-GNN}

Suppose the example graph adjacency matrix is ${\scriptsize \left[  \begin{smallmatrix} 1&0&1&0\\0&1&0&1\\1&0&0&1\\0&1&1&0 \end{smallmatrix} \right]}$, the conventional GNN Block of forward and backward pass~\cite{bai2019deepequilibriummodels} 
performs product full-dimensional node embedding or gradient processed by Rectified Linear Units (ReLU~\cite{agarap2019deeplearningusingrectified}) $ReLU(X^{l})$ in ${\scriptsize\left[ \begin{smallmatrix} 
0.37 & 0.76 & 0 & 0.79 \\ 
0.58 & 0 & 0.57 & 0 \\ 
0 & 0.55 & 0.95 & 0 \\ 
0 & 0 & 0.23 & 0 
\end{smallmatrix} \right]}$ and $ReLU(\frac{dX^{l+1}}{dL})$ in ${\scriptsize\left[ \begin{smallmatrix} 
0.23 & 0 & 0.89 & 0 \\ 
0.45 & 0.78 & 0 & 0.67 \\ 
0 & 0 & 0 & 0.88 \\ 
0 & 0.33 & 0.66 & 0 
\end{smallmatrix} \right]}$, resulting in ${\scriptsize\left[\begin{smallmatrix} 
0.37 & 1.31 & 0.95 & 0.79 \\ 
0.58 & 0 & 0.8 & 0 \\ 
0.37 & 0.76 & 0.23 & 0.79 \\ 
0.58 & 0.55 & 1.52 & 0 
\end{smallmatrix} \right]}$ and ${\scriptsize\left[ \begin{smallmatrix} 
0.23 & 0 & 0.89 & 0.38 \\ 
0.56 & 1.11 & 0.66 & 0.67 \\ 
0.23 & 0.33 & 1.55 & 0 \\ 
0.45 & 1.11 & 0 & 1.05 
\end{smallmatrix} \right]}$ as $X^{l+1}$ and $\frac{dX^{l}}{dL}$ respectively. In contrast, our GSR-GNN Block introduces Group Sparse $\text{GS}(X, k) \rightarrow (V, I)$ as a counterpart to ReLU. Where values $V \in \mathbb{R}^{N \times k}$ and indices $I \in \mathbb{Z}^{N \times k}$ are equivalent to sparsifying the node embedding with pre-defined $k$ values, which is  $\text{Val}$:${\scriptsize\left[ \begin{smallmatrix} 0.79 \\ 0.58 \\ 0.95 \\ 0.86 \end{smallmatrix} \right]}$ and $\text{Idx}$:${\scriptsize\left[ \begin{smallmatrix} 3 \\ 0 \\ 2 \\ 3 \end{smallmatrix} \right]}$ for forward block example, where each row of $\text{val}$ is the K-most significant elements selected from full-dimentional node embedding. And $\text{Val}$:${\scriptsize\left[ \begin{smallmatrix} 0.12 \\ 0.45 \\ 0.55 \\ 0.99 \end{smallmatrix} \right]}$ for the backward block example while keeping the same $\text{val}$ saved since forward. The indices $I/\text{idx}$ record the "activated" neurons' positions to source $V/\text{val}$ in the embedding from the forward pass. It is preserved and used in the backward process to source the gradient that shares the same positions as in the embedding; this step is an indispensable contributing factor to performance. Then, our sparse message passing occurs via:
{\begin{align}
X^{l+1} &= \text{GSR-GNN-Forward-Block}(V^l, I^l, G) \\
\frac{\partial \mathcal{L}}{\partial X^l} &= \text{GSR-GNN-Backward-Block}(M^l, I^l, G)
\end{align}}
Then, $X^{l+1}$ is $\scriptsize{\left[  \begin{smallmatrix} 
0 & 0 & 0.95 & 0.79 \\ 
0.58 & 0 & 0 & 0.86 \\ 
0 & 0 & 0 & 1.65 \\ 
0.58 & 0 & 0.95 & 0 
\end{smallmatrix} \right]}$ and $\frac{\partial \mathcal{L}}{\partial X^l}$ becomes ${\scriptsize\left[ \begin{smallmatrix} 
0 & 0 & 0.67 & 0 \\ 
1.44 & 0 & 0 & 0 \\ 
0 & 0 & 1.11 & 0 \\ 
0 & 0 & 0 & 1.0 
\end{smallmatrix} \right]}$ through product and scatter.
With $(V, I)$, our forward block promptly uses values in the node embedding indexed by the indices while skipping zero elements, rather than accessing full-dimensional embeddings per node when it comes to the partially reversible backward phase. Similarly, the GSR-GNN backward block also efficiently utilizes $I$ generated in the forward block to record neurons' positions for reference, so the corresponding backward block only reads the gradients whose positions are recorded by $I$ to compute and then finish one learning step.

\begin{algorithm}[t]
\scriptsize
\label{alg:gsr-gnn-forward}
\caption{GSR-GNN Forward}\label{alg:gsr-gnn-forward}
\begin{algorithmic}[1]
\State \textbf{Input:} Graph $G$, features $X^{0}$, layers $L$, GS parameter $k$
\State \textbf{Output:} Predictions $\hat{y}$

\For{$l=1$ to $L$}
  \State $X_1^l, X_2^l \gets \operatorname{Split}(X^{l-1})$; 
  \State $V_1^l, I_1^l \gets \operatorname{GS}(X_1^l, k)$ \Comment{Group partition \& GS}
  \State $M_1^l \gets \operatorname{GSR\text{-}Forward}(V_1^l, I_1^l, G)$
  \State $X_2^{l+1} \gets X_2^l + M_1^l$
  \State $V_2^l, I_2^l \gets \operatorname{GS}(X_2^{l+1}, k)$
  \State $M_2^l \gets \operatorname{GSR\text{-}Forward}(V_2^l, I_2^l, G)$
  \State $X_1^{l+1} \gets \operatorname{scatter}(V_1^l, I_1^l) + M_2^l$
  \State $X^{l+1} \gets \operatorname{Concat}(X_1^{l+1}, X_2^{l+1})$ \Comment{Stack groups}
\EndFor

\State \Return $\operatorname{OutputLayer}(X^{L})$
\end{algorithmic}
\end{algorithm}

\subsection{Enhanced Memory Management}
Our GSR-GNN framework changes the in-group workflow compared to conventional RevGNN. Unlike RevGNN's frequent detach-copy-delete cycles for both node embeddings (forward pass) and gradient tensors (backward pass), our custom blocks eliminate these costly memory operations.

This optimization yields three key benefits: (1) \textbf{reduced fragmentation} by avoiding unnecessary tensor copies, (2) \textbf{improved memory utilization} through elimination of inactive memory pools, and (3) \textbf{selective save-for-backward} operations that only store indices when required for subsequent backward blocks.

While our approach maintains $O(LNK)$ memory complexity linear in the number of layers $L$, the actual memory footprint remains significantly lower than RevGNN within the practical depth range. This design ensures predictable memory usage while maximizing GPU memory efficiency for deep graph neural networks.

\subsection{Forward Phase of GSR-GNN}
Figure \ref{fig:GSR-GNN-Fwd} and \ref{fig:GSR-GNN-Arch-Bckwd} provide an overview of GSR-GNN. For each layer, during the forward propagation shown in Figure \ref{fig:GSR-GNN-Fwd}, the input node embedding is partitioned evenly into two groups, $\text{split}
{\scriptsize \left[  \begin{smallmatrix} 
0.37 & 0.76 & 0.45 & 0.17 \\ 
0.58 & 0.31 & 0.21 & 0.40 \\ 
0.92 & 0.55 & 0.73 & 0.07 \\ 
0.14 & 0.02 & 0.15 & 0.49 
\end{smallmatrix} \right]}
=$
{\setlength{\fboxsep}{0pt}\colorbox[HTML]{E5CCFF}{$
{\scriptsize\left[ \begin{smallmatrix} 
0.37 & 0.76 \\ 
0.58 & 0.31 \\ 
0.92 & 0.55 \\ 
0.14 & 0.02 
\end{smallmatrix} \right]}$}}
{\setlength{\fboxsep}{0pt}\colorbox[HTML]{D5E8D4}{${\scriptsize\left[ \begin{smallmatrix} 
0.45 & 0.17 \\ 
0.21 & 0.40 \\ 
0.73 & 0.07 \\ 
0.15 & 0.49 
\end{smallmatrix} \right]}$}}
. This is step \textcircled{1} from line 5, Algorithm \ref{alg:gsr-gnn-forward}. Group 1 {\setlength{\fboxsep}{0pt}\colorbox[HTML]{D5E8D4}{${\scriptsize\left[ \begin{smallmatrix} 
0.45 & 0.17 \\ 
0.21 & 0.40 \\ 
0.73 & 0.07 \\ 
0.15 & 0.49 
\end{smallmatrix} \right]}$}}
firstly proceeds through the GS-Op at \textcircled{2} (line 5, Algorithm\ref{alg:gsr-gnn-forward}) to produce the compressed CBSR-format Val: {\setlength{\fboxsep}{0pt}\colorbox[HTML]{D5E8D4}{$ \scriptsize{\left[ \begin{smallmatrix} 0.45 \\ 0.40 \\ 0.73 \\ 0.49 \end{smallmatrix} \right]}$}}, idx: {\setlength{\fboxsep}{0pt}\colorbox[HTML]{D5E8D4}{$ \scriptsize{\left[ \begin{smallmatrix} 0 \\ 1 \\ 0 \\ 1 \end{smallmatrix} \right] }$}}, which become the input of the GSR-GNN forward block first (\textcircled{3}, line 6 of Algorithm \ref{alg:gsr-gnn-forward}), then the result {\setlength{\fboxsep}{0pt}\colorbox[HTML]{D5E8D4}{$\scriptsize{\left[\begin{smallmatrix} 
1.18 & 0 \\ 
0 & 0.89 \\ 
0.94 & 0 \\ 
0 & 1.13 
\end{smallmatrix} \right] }$} }added to Group 2 {\setlength{\fboxsep}{0pt}\colorbox[HTML]{E5CCFF}{$\scriptsize{\left[\begin{smallmatrix} 
0.45 & 0.17 \\ 
0.21 & 0.40 \\ 
0.73 & 0.07 \\ 
0.15 & 0.49 
\end{smallmatrix} \right]}$}}, which subsequently enters the second GS-Op at \textcircled{5} at line 8 and produces Val:{\setlength{\fboxsep}{0pt}\colorbox[HTML]{E5CCFF}{$\scriptsize{\left[ \begin{smallmatrix} 1.55 \\ 1.2 \\ 1.86 \\ 1.15 \end{smallmatrix} \right]}$}} and Idx: {\setlength{\fboxsep}{0pt}\colorbox[HTML]{E5CCFF}{$\scriptsize{\left[ \begin{smallmatrix} 0 \\ 1 \\ 0 \\ 1 \end{smallmatrix} \right]}$}}
, then another forward block at \textcircled{6}, line 9.
Following this, the output of the second forward block {\setlength{\fboxsep}{0pt}\colorbox[HTML]{E5CCFF}{$\scriptsize{\left[ \begin{smallmatrix} 
3.35 & 0 \\ 
0 & 2.35 \\ 
2.65 & 0 \\ 
0 & 3.06 
\end{smallmatrix} \right]}$}} is also added back to group 1 at \textcircled{7} to be the updated channel 1, shown by line 10, 11 of Algorithm\ref{alg:gsr-gnn-forward}, meanwhile becomes the updated channel 2 of the node embedding at \textcircled{8} with the updated group 1, $X_1^{l+1}=$ {\setlength{\fboxsep}{0pt}\colorbox[HTML]{D5E8D4}{$\scriptsize{\left[ \begin{smallmatrix} 
3.8 & 0 \\ 
0 & 2.75 \\ 
3.38 & 0 \\ 
0 & 3.55 
\end{smallmatrix} \right]}$}}, this also updates channel 2 of the node embedding to be passed to the next layer. 

\begin{algorithm}[t]
\caption{GSR-GNN Backward}
\label{alg:gsr-gnn-backward}
\scriptsize
\begin{algorithmic}[1]

\State \textbf{Input:} Gradient $\partial L / \partial \hat{y}$ \quad \textbf{Output:} Gradient $\partial L / \partial X^{0}$

\State $g^{L} \gets \text{OutputLayerBackward}(\partial L / \partial \hat{y})$

\For{$l = L$ down to $1$}
    \State $(g_1^{l+1}, g_2^{l+1}) \gets \text{Split}(g^{l+1})$; \quad $(V_2, I_2) \gets \text{GS}(g_2^{l+1}, k)$ \Comment{ Group partition \& GS}
    
    \State $M_1 \gets g_1^{l+1} - \text{GSRBlock}(V_2, I_2, G)$
    
    \State $(V_1, I_1) \gets \text{GS}(M_1, k)$; \quad $Z_1 \gets \text{GSRBlock}(V_1, I_1, G)$
    
    \State $M_2 \gets V_2.\text{scatter}(I_2) - Z_1$
    
    \State $g_1^{l} \gets \text{GSRBlockBackward}(M_1, I_1, G)$; \quad $g_2^{l} \gets \text{GSRBlockBackward}(M_2, I_2, G)$ 
    
    \State $g^{l} \gets \text{Concat}(g_1^{l}, g_2^{l})$
\EndFor

\State \textbf{Return} $g^{0}$
\end{algorithmic}

\end{algorithm}
\vspace{-0.3cm}


    
    
    



\subsection{Backward Phase of GSR-GNN}
During the backward propagation in Figure \ref{fig:GSR-GNN-Arch-Bckwd} and Algorithm \ref{alg:gsr-gnn-backward}, the gradient passed from the next layer is split into two groups $\text{split}
{\scriptsize\left[ \begin{smallmatrix} 
0.37 & -0.82 & 0.15 & -0.53 \\ 
-0.21 & 0.68 & -0.94 & 0.07 \\ 
0.91 & -0.13 & 0.45 & -0.76 \\ 
-0.62 & 0.29 & -0.08 & 0.83 
\end{smallmatrix} \right]}$
$=$
{\setlength{\fboxsep}{0pt}\colorbox[HTML]{E5CCFF}{${\scriptsize\left[ \begin{smallmatrix} 
0.37 & -0.82 \\ 
-0.21 & 0.68 \\ 
0.91 & 0.13 \\ 
-0.62 & 0.29 
\end{smallmatrix} \right]}$}},
{\setlength{\fboxsep}{0pt}\colorbox[HTML]{D5E8D4}{${\scriptsize\left[ \begin{smallmatrix} 
0.15 & -0.53 \\ 
-0.94 & 0.07 \\ 
0.45 & -0.76 \\ 
-0.08 & 0.83 
\end{smallmatrix} \right]}$}}
 (step \textcircled{1}, line 4, Algorithm\ref{alg:gsr-gnn-backward}), whereas, at this time, group 2 {\setlength{\fboxsep}{0pt}\colorbox[HTML]{E5CCFF}{${\scriptsize\left[ \begin{smallmatrix} 
0.37 & -0.82 \\ 
-0.21 & 0.68 \\ 
0.91 & 0.13 \\ 
-0.62 & 0.29 
\end{smallmatrix} \right]}$}} goes through GS-Op to output CBSR values {\setlength{\fboxsep}{0pt}\colorbox[HTML]{E5CCFF}{$\scriptsize{\left[\begin{smallmatrix} 0.37 \\ 0.68 \\ 0.91 \\ 0.29 \end{smallmatrix} \right]}$}} and indices {\setlength{\fboxsep}{0pt}\colorbox[HTML]{E5CCFF}{$\scriptsize{\left[ \begin{smallmatrix} 0 \\ 1 \\ 0 \\ 1 \end{smallmatrix} \right]}$}} at \textcircled{2}, and enter the GSR-GNN forward block at \textcircled{3} at line 5 in Algorithm \ref{alg:gsr-gnn-backward} and note that positions (indices) of the activated neurons are saved for the step at line 8. Then, the output of the first forward block  {\setlength{\fboxsep}{0pt}\colorbox[HTML]{D5E8D4}{$\scriptsize{\left[ \begin{smallmatrix} 
1.28 & 0 \\ 
0 & 0.97 \\ 
0.66 & 0 \\ 
0 & 1.59 
\end{smallmatrix} \right]}$}} goes to the subtraction with group 1 (line 7, Alg \ref{alg:gsr-gnn-backward}), the result  {\setlength{\fboxsep}{0pt}\colorbox[HTML]{D5E8D4}{$\scriptsize{\left[ \begin{smallmatrix} 
1.13 & 0.53 \\ 
0.94 & 0.9 \\ 
0.21 & 0.76 \\ 
0.08 & 0.76 
\end{smallmatrix} \right]}$}} goes into two ways: one enters another GS-Op at \textcircled{4} with the result {\setlength{\fboxsep}{0pt}\colorbox[HTML]{D5E8D4}{$\scriptsize{\left[\begin{smallmatrix} -0.53 \\ -0.9\\ -0.21 \\ -0.08 \end{smallmatrix} \right]}$}} and indices {\setlength{\fboxsep}{0pt}\colorbox[HTML]{D5E8D4}{$\scriptsize{\left[ \begin{smallmatrix} 1 \\ 1 \\ 0 \\ 0 \end{smallmatrix} \right]}$}}, which enter the second forward block at \textcircled{5}, whose result {\setlength{\fboxsep}{0pt}\colorbox[HTML]{D5E8D4}{$\scriptsize\left[ \begin{smallmatrix} 
0 & -0.74 \\ 
0 & -0.98 \\ 
-0.61 & 0 \\ 
-1.11 & 0 
\end{smallmatrix} \right]$ }}performs subtraction with group 2 at \textcircled{6} to be the input of its GSR-GNN backward block at \blackcircle{1}; the other is kept for the dedicated backward block at \blackcircle{2}. This process is covered by lines 8 in Algorithm \ref{alg:gsr-gnn-backward}. Using the remembered positions from the previous forward-style recomputation, both channels of the gradient $\frac{\partial L}{\partial X_1^{l}}$ {\setlength{\fboxsep}{0pt}\colorbox[HTML]{E5CCFF}{$\scriptsize{\left[ \begin{smallmatrix} 
1.89 & 0 \\ 
0 & 1.95 \\ 
0.66 & 0 \\ 
0 & 3.18 
\end{smallmatrix} \right]}$}} and $\frac{\partial L}{\partial X_2^{l}}${\setlength{\fboxsep}{0pt}\colorbox[HTML]{D5E8D4}{ $\scriptsize{\left[ \begin{smallmatrix} 
0 & -0.74 \\ 
0 & -0.98 \\ 
-0.61 & 0 \\ 
-1.11 & 0 
\end{smallmatrix} \right]}$}} are updated to be passed to the last layer at \blackcircle{3}.

\section{Experiments}

This section presents our experimental setup and evaluates the performance of GSR-GNN against baseline methods. We demonstrate the framework's advantages in terms of training acceleration and memory efficiency while maintaining comparable accuracy.

\begin{table}
\footnotesize
\centering
\caption{Circuit Graph Dataset Summary (Ordered by Size)}
\label{tab:compact_graph_summary}
\renewcommand{\arraystretch}{0.75}
\setlength{\tabcolsep}{1.2pt} 
\begin{tabular}{lclcccccp{0.8cm}} 
\toprule
\textbf{Size} & \textbf{\#} & \textbf{Name} & \textbf{Node} & \textbf{Edge} & \textbf{Train/Val/Test} & \textbf{Labels} & \textbf{Range} \\ 
\midrule
\multirow{2}{*}{L} 
    & 1 & 5324-F & 69.8k & 22.5M & 48.8/10.5/10.5k & 131 & [0.0, 1.0] \\ 
    & 2 & 4474-F & 65.3k & 21.8M & 52.3/6.5/6.5k & 255 & [0.0, 1.04] \\
\midrule
\multirow{5}{*}{M} 
    & 3 & 876-R & 49.6k & 12.2M & 39.7/5.0/5.0k & 60 & [0.0, 0.22] \\
    & 4 & 1522-R & 49.5k & 12.2M & 39.6/5.0/5.0k & 106 & [0.0, 0.36] \\
    & 5 & 1886-R & 49.1k & 11.7M & 39.3/4.9/4.9k & 134 & [0.0, 0.58] \\
    & 6 & 639-R & 49.1k & 11.7M & 34.3/7.4/7.4k & 105 & [0.0, 0.48] \\
    & 7 & 356-R & 49.0k & 11.7M & 39.2/4.9/4.9k & 271 & [0.0, 0.90] \\
\midrule
\multirow{3}{*}{S} 
    & 8 & 6859-F & 44.3k & 2.7M & 35.4/4.4/4.4k & 28 & [0.0, 1.0] \\
    & 9 & 6227-F & 43.4k & 2.7M & 34.7/4.3/4.3k & 148 & [0.0, 3.75] \\
    & 10 & 7157-R & 38.7k & 6.8M & 30.9/3.9/3.9k & 153 & [0.0, 0.66] \\
\bottomrule

\end{tabular}
\begin{tablenotes}
\small
\item 
Note: "F" and "R" in the column Name refer to "FPU" and "RISC-Y".
\end{tablenotes}
\vspace{-0.4cm}
\end{table}


\noindent\textbf{Platform:} All experiments were conducted on an AMD EPYC 7763 64-Core Processor server (504GB RAM) and a single NVIDIA A6000-48GB GPU (CUDA toolkit 12.6), PyG version 2.5.0, Pytorch 2.1.2. 

\noindent\textbf{Dataset:} We evaluate our approach using the CircuitNet dataset~\cite{10158384}, which originates from over 10,000 circuit graph samples serving various downstream tasks. We randomly sample 10 representative circuit designs (detailed in Table \ref{tab:compact_graph_summary}) following the provided pre-processing method without sub-graph partitioning to ensure design size variability and fully expose the scale of graph data.
We focus primarily on congestion prediction, a critical step in the circuit design process~\cite{jiang2024circuitnet, 10158384, 9598835}. Following standard practice in circuit design applications~\cite{yang2022versatile}, we evaluate performance using three rank correlation metrics: Pearson, Kendall, and Spearman correlation scores, with higher values indicating better performance.

\noindent\textbf{Models and Configuration:}
We compared our model with state-of-the-art GraphSAGE~\cite{liu2020graphsage}, GCN~\cite{gcn}, and the official RevGNN~\cite{revgnn}. The hyperparameters include: hidden dimension, varying between 256 and 384; the number of layers, ranging from 10 to 100 with a step of ten layers each; While ReLU from baselines doesn't need parameter choices, the K-values for our GS-Op fall in the set $\{4,8,16,32,64\}$, the training takes up 250 epochs, the train, validation and test set division are given in Table \ref{tab:compact_graph_summary}.

\begin{figure}[t]
    \centering
    \includegraphics[width=\linewidth]{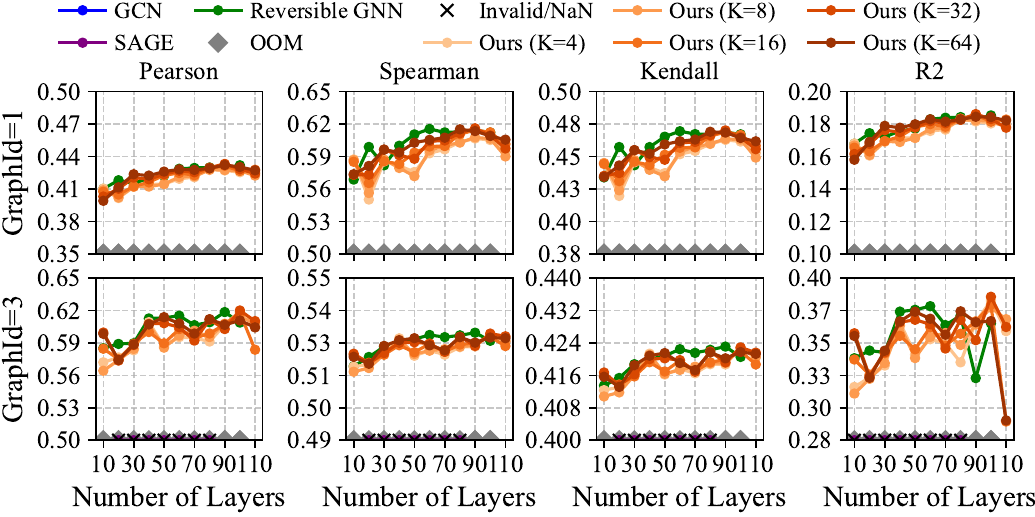}
    \caption{Correlation performance of GSR-GNN over GCN/SAGE/RevGNN in selected graphs (Graph 1 and Graph 3 in Table \ref{tab:compact_graph_summary}). The results are averaged over hidden dimension=256 and 384. 
    The higher correlation scores mean better performance, non-reversible baselines (SAGE, GCN) are out-of-range due to very low scores or out-of-memory errors.
    }
    \vspace{-0.1in}
    \label{fig:approximate-accuracy}
\end{figure}

\begin{figure}[htbp]
    \centering
    \includegraphics[width=1\linewidth]{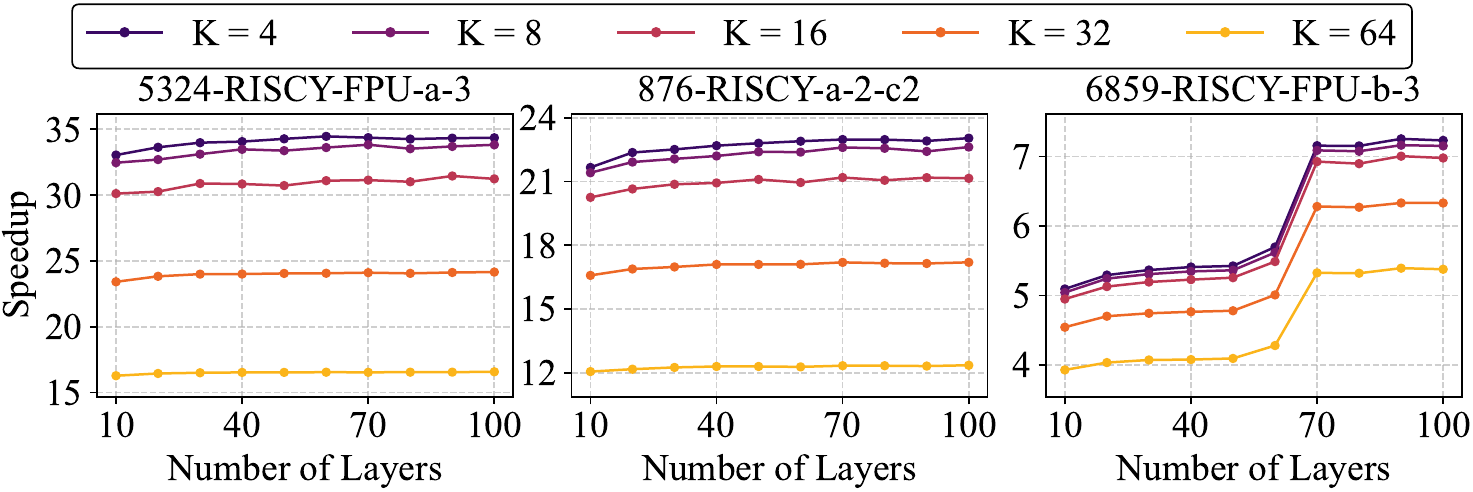}
    \caption{Speedup of GSR-GNN over Reversible GNNs at varying sparsity levels defined by $K$, the greater the $K$, the more elements preserved per row of activations and lower sparsity.}
    \vspace{-0.2in}
    \label{fig:speed-up-ratio}
\end{figure}

\subsection{Experimental Results}

This section demonstrates that GSR-GNN achieves training acceleration while maintaining comparable correlation performance to standard reversible deep GNN models. Given that non-reversible GCN and SAGE models exhibit severe performance fluctuations with increasing depth and remain inferior to RevGNN (Figure  \ref{fig:revgnn-perf-comp}), our primary comparison focuses on RevGNN.

Figure \ref{fig:approximate-accuracy} evaluates GSR-GNN with GS-Op $(K\in\{4,8,16,32,64\})$ against RevGNN on CircuitNet for congestion prediction across 10 randomly sampled graphs in Table~\ref{tab:compact_graph_summary}. The correlation metrics demonstrate that GSR-GNN maintains performance comparable to RevGNN, with mean differences of +0.0026 (Pearson), +0.0067 (Spearman), +0.0053 (Kendall), and -0.023 ($R^2$). These negligible differences indicate that GS-Op preserves model accuracy while enabling the observed speedups. 

Figure \ref{fig:speed-up-ratio} illustrates the end-to-end training speedup of GSR-GNN at two hidden channels, 256 and 384, over RevGNN across three circuit scales.  GSR-GNN achieves over $35\times$ (K=4, 384 hidden channels) for graph 5324-RISCY-FPU-a-3, over $4\times$ (K=64) for graph 6859-RISCY-FPU-b-3, with performance inversely correlated to sparsity level.  In addition, we revisit the breakdown of training step runtime since Figure \ref{fig:revgnn-memory-time}, comparing our GSR-GNN against RevGNN on the same graph dataset 356-RISCY-a1-c5. 

Table \ref{tab:vertical_distribution} shows that GSR-GNN has obvious acceleration across three sections: Forward, Backward, and cudaMemCpy. The most significant training component, Backward, is reduced to 1.52 seconds at best, and forward is 0.36 seconds. Our memory management optimization consistently cuts cudaMemcpy time to ~0.0001 seconds during the whole training workflow, making a final best speedup of $18.4\times$. Though increasing sparsity in terms of $K$ lowers the relative speedup effect, the worst end-to-end speedup can still reach $9.1\times$.



\begin{table}[t]
\footnotesize
\centering
\caption{Runtime Distribution: 100-layer, 384-hidden channels Baseline (RevGNN) vs Ours (Graph 3 in Table \ref{tab:compact_graph_summary})}
\label{tab:vertical_distribution}
\setlength{\tabcolsep}{3pt}
\begin{tabular}{lcccc} 
\toprule 
\textbf{Method} & \textbf{Forward$\downarrow$} & \textbf{Backward$\downarrow$} & \textbf{cudaMemCpy$\downarrow$} & \textbf{Total/Spdup$\uparrow$} \\
\midrule 
Baseline & 5.84s (20.8\%) & 20.09s (71.6\%) & 2.13s (7.6\%) & 28.06s  \\
\midrule 
Ours (K=4) & 0.36s (23.7\%) & 1.16s (76.2\%) & 0.001s (0.1\%) & 1.52s (18.4×) \\
Ours (K=8) & 0.37s (24.2\%) & 1.17s (75.7\%) & 0.001s (0.1\%) & 1.54s (18.2×) \\
Ours (K=16) & 0.40s (24.1\%) & 1.25s (75.9\%) & 0.001s (0.1\%) & 1.64s (17.1×) \\
Ours (K=32) & 0.50s (24.1\%) & 1.59s (75.9\%) & 0.001s (0.1\%) & 2.09s (13.4×) \\
Ours (K=64) & 0.76s (24.7\%) & 2.31s (75.2\%) & 0.001s (0.1\%) & 3.07s (9.1×) \\
\bottomrule 
\end{tabular}
\vspace{-15pt}
\end{table}

\begin{figure}[htbp]
    \centering
    \begin{subfigure}[b]{0.95\linewidth}
    \includegraphics[width=\linewidth]{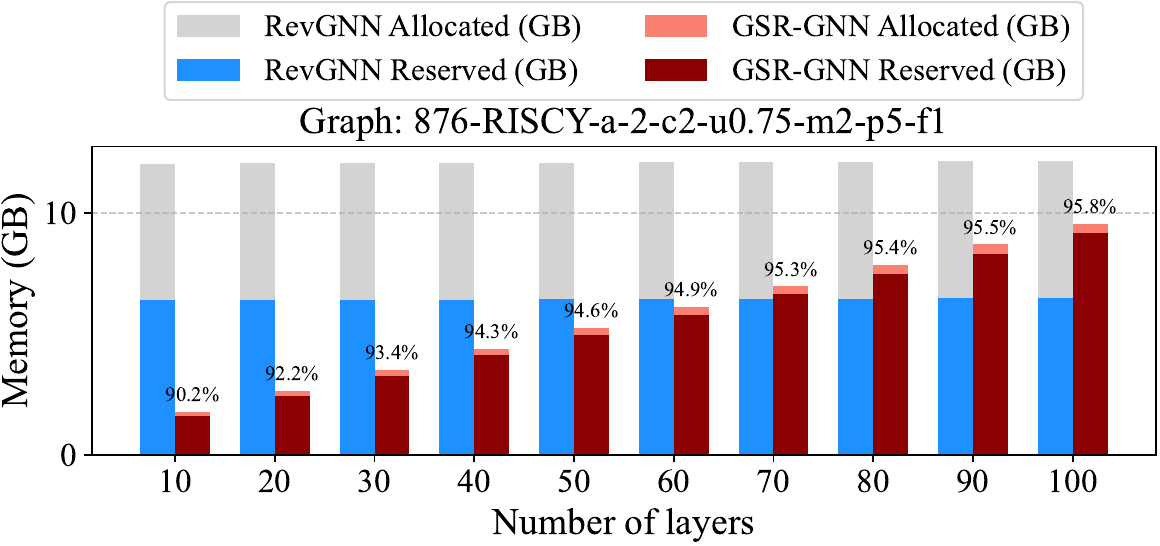}
    \caption{876-RISCY-a-2-c2-u0.75-m2-p5-f1}
    \label{fig:876}
    \end{subfigure}
    \hfill
    
    \begin{subfigure}[b]{0.95\linewidth}
    \includegraphics[width=\linewidth]{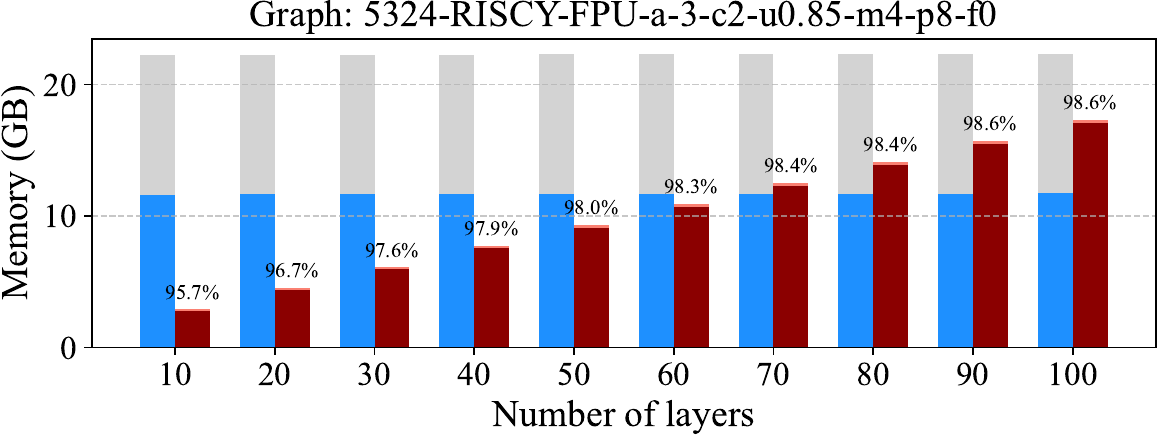}
    \caption{5324-RISCY-FPU-a-3-c2-u0.85-m4-p8-f0}
    \label{fig:5324}
    \end{subfigure}
    \caption{Memory efficiency comparison between RevGNN and GSN-GNN on 2 sample graphs, graph 1 and graph 3 in Table \ref{tab:compact_graph_summary}. The memory efficiency, expressed as a percentage, is the ratio of Allocated GPU Memory to Reserved GPU Memory during the training phase of models.}
    \vspace{-0.1in}
    \label{fig:mem}
\end{figure}

Furthermore, Figure 
\ref{fig:mem} 
presents the memory efficiency of GSR-GNN compared to RevGNN on graph 876-R and 5324-R from Table \ref{tab:compact_graph_summary}. Due to optimized memory management and compressed node embeddings, within the optimal range of the number of layers, GSR-GNN consistently consumes less GPU memory than RevGNN, even at 100 layers. GSR-GNN comes with a clear, minimum gap of on average 2.58GB and 5.043GB, respectively, for both graphs. Although the memory savings decrease with increasing layers, GSR-GNN consistently maintains over 90\% memory utilization, whereas RevGNN always exhibits 
approximately 50\% memory waste.
This shows that our GSR-GNN reduces the maximum number of GNNs that can be trained on circuit graph datasets of various sizes. 
Additionally, this enables training of circuit graph datasets on GPUs with less than 8 GB of memory at layers less than 30.  To be closer to baseline performance while maintaining relative speedup, one can also increase both layers and lower the sparsity to achieve near-optimal performance in correlation metrics.



\section{Conclusion}
In our paper, we propose a GNN training framework targeting faster and more GPU memory-efficient deep GNN training on circuit graph datasets. We redesign the grouped convolution and reversible residual module of RevGNN with custom grouped sparse reversible GNN blocks and applied it to graph datasets sampled from CircuitNet in varied, representative sizes. In the evaluation on target circuit graph datasets, our framework achieves apparent training acceleration effect (over $35\times$ at most), lower GPU memory reservation ($87.22\%$ GPU memory saving at most), and higher efficiency (98.6\% rate of active memory over total reversed memory), with almost unchanged performance in correlation metrics from the congestion prediction task.
\begin{acks}
This research was supported in part by NSF SHF-2505770.
\end{acks}

\bibliographystyle{ACM-Reference-Format}
\bibliography{sample-base}

@INPROCEEDINGS{7459464,
  author={Sayed-Ahmed, Amr and Große, Daniel and Kühne, Ulrich and Soeken, Mathias and Drechsler, Rolf},
  booktitle={2016 Design, Automation \& Test in Europe Conference \& Exhibition (DATE)}, 
  title={Formal verification of integer multipliers by combining Gröbner basis with logic reduction}, 
  year={2016},
  volume={},
  number={},
  pages={1048-1053},
  doi={}}

@ARTICLE{8105885,
  author={Yu, Cunxi and Ciesielski, Maciej and Mishchenko, Alan},
  journal={IEEE Transactions on Computer-Aided Design of Integrated Circuits and Systems}, 
  title={Fast Algebraic Rewriting Based on And-Inverter Graphs}, 
  year={2018},
  volume={37},
  number={9},
  pages={1907-1911},
  doi={10.1109/TCAD.2017.2772854}}

@article{Kaufmann+2022+285+291,
url = {https://doi.org/10.1515/itit-2022-0039},
title = {Formal verification of multiplier circuits using computer algebra},
title = {},
author = {Daniela Kaufmann},
pages = {285--291},
volume = {64},
number = {6},
journal = {it - Information Technology},
doi = {doi:10.1515/itit-2022-0039},
year = {2022},
lastchecked = {2023-05-20}
}

@inproceedings{li2019deepgcns,
  title={DeepGCNs: Can GCNs go as deep as CNNs?},
  author={Li, Guohao and Muller, Matthias and Thabet, Ali and Ghanem, Bernard},
  booktitle={Proceedings of the IEEE/CVF international conference on computer vision},
  pages={9267--9276},
  year={2019}
}

@article{li2020deepergcn,
  title={Deepergcn: All you need to train deeper gcns},
  author={Li, Guohao and Xiong, Chenxin and Thabet, Ali and Ghanem, Bernard},
  journal={arXiv preprint arXiv:2006.07739},
  year={2020}
}

@article{gomez2017reversible,
  title={The reversible residual network: Backpropagation without storing activations},
  author={Gomez, Aidan N and Ren, Mengye and Urtasun, Raquel and Grosse, Roger B},
  journal={Advances in neural information processing systems},
  volume={30},
  year={2017}
}

@misc{revgnn,
      title={Training Graph Neural Networks with 1000 Layers}, 
      author={Guohao Li and Matthias Müller and Bernard Ghanem and Vladlen Koltun},
      year={2022},
      eprint={2106.07476},
      archivePrefix={arXiv},
      primaryClass={cs.LG},
      url={https://arxiv.org/abs/2106.07476}, 
}

@article{hu2020open,
  title={Open graph benchmark: Datasets for machine learning on graphs},
  author={Hu, Weihua and Fey, Matthias and Zitnik, Marinka and Dong, Yuxiao and Ren, Hongyu and Liu, Bowen and Catasta, Michele and Leskovec, Jure},
  journal={Advances in neural information processing systems},
  volume={33},
  pages={22118--22133},
  year={2020}
}

@ARTICLE{9598835,
  author={Rapp, Martin and Amrouch, Hussam and Lin, Yibo and Yu, Bei and Pan, David Z. and Wolf, Marilyn and Henkel, Jörg},
  journal={IEEE Transactions on Computer-Aided Design of Integrated Circuits and Systems}, 
  title={MLCAD: A Survey of Research in Machine Learning for CAD Keynote Paper}, 
  year={2022},
  volume={41},
  number={10},
  pages={3162-3181},
  doi={10.1109/TCAD.2021.3124762}}

@misc{chowdhury2021openabcdlargescaledatasetmachine,
      title={OpenABC-D: A Large-Scale Dataset For Machine Learning Guided Integrated Circuit Synthesis}, 
      author={Animesh Basak Chowdhury and Benjamin Tan and Ramesh Karri and Siddharth Garg},
      year={2021},
      eprint={2110.11292},
      archivePrefix={arXiv},
      primaryClass={cs.LG},
      url={https://arxiv.org/abs/2110.11292}, 
}

@ARTICLE{10158384,
  author={Chai, Zhuomin and Zhao, Yuxiang and Liu, Wei and Lin, Yibo and Wang, Runsheng and Huang, Ru},
  journal={IEEE Transactions on Computer-Aided Design of Integrated Circuits and Systems}, 
  title={CircuitNet: An Open-Source Dataset for Machine Learning in VLSI CAD Applications With Improved Domain-Specific Evaluation Metric and Learning Strategies}, 
  year={2023},
  volume={42},
  number={12},
  pages={5034-5047},
  doi={10.1109/TCAD.2023.3287970}}

@inproceedings{
jiang2024circuitnet,
title={CircuitNet 2.0: An Advanced Dataset for Promoting Machine Learning Innovations in Realistic Chip Design Environment},
author={Xun Jiang and zhuomin chai and Yuxiang Zhao and Yibo Lin and Runsheng Wang and Ru Huang},
booktitle={The Twelfth International Conference on Learning Representations},
year={2024},
url={https://openreview.net/forum?id=nMFSUjxMIl}
}

@misc{dinh2015nicenonlinearindependentcomponents,
      title={NICE: Non-linear Independent Components Estimation}, 
      author={Laurent Dinh and David Krueger and Yoshua Bengio},
      year={2015},
      eprint={1410.8516},
      archivePrefix={arXiv},
      primaryClass={cs.LG},
      url={https://arxiv.org/abs/1410.8516}, 
}

@inproceedings{10.1145/3489517.3530597,
author = {Guo, Zizheng and Liu, Mingjie and Gu, Jiaqi and Zhang, Shuhan and Pan, David Z. and Lin, Yibo},
title = {A timing engine inspired graph neural network model for pre-routing slack prediction},
year = {2022},
isbn = {9781450391429},
publisher = {Association for Computing Machinery},
address = {New York, NY, USA},
url = {https://doi.org/10.1145/3489517.3530597},
doi = {10.1145/3489517.3530597},
booktitle = {Proceedings of the 59th ACM/IEEE Design Automation Conference},
pages = {1207–1212},
numpages = {6},
location = {San Francisco, California},
series = {DAC '22}
}

@misc{zhang2021evaluatingdeepgraphneural,
      title={Evaluating Deep Graph Neural Networks}, 
      author={Wentao Zhang and Zeang Sheng and Yuezihan Jiang and Yikuan Xia and Jun Gao and Zhi Yang and Bin Cui},
      year={2021},
      eprint={2108.00955},
      archivePrefix={arXiv},
      primaryClass={cs.LG},
      url={https://arxiv.org/abs/2108.00955}, 
}

@misc{agarap2019deeplearningusingrectified,
      title={Deep Learning using Rectified Linear Units (ReLU)}, 
      author={Abien Fred Agarap},
      year={2019},
      eprint={1803.08375},
      archivePrefix={arXiv},
      primaryClass={cs.NE},
      url={https://arxiv.org/abs/1803.08375}, 
}

@article{Albert_2002,
   title={Statistical mechanics of complex networks},
   volume={74},
   ISSN={1539-0756},
   url={http://dx.doi.org/10.1103/RevModPhys.74.47},
   DOI={10.1103/revmodphys.74.47},
   number={1},
   journal={Reviews of Modern Physics},
   publisher={American Physical Society (APS)},
   author={Albert, Réka and Barabási, Albert-László},
   year={2002},
   month=jan, pages={47–97} }

@inproceedings{
yang2022versatile,
title={Versatile Multi-stage Graph Neural Network for Circuit Representation},
author={Shuwen Yang and Zhihao Yang and Dong Li and Yingxue Zhang and Zhanguang Zhang and Guojie Song and Jianye HAO},
booktitle={Advances in Neural Information Processing Systems},
editor={Alice H. Oh and Alekh Agarwal and Danielle Belgrave and Kyunghyun Cho},
year={2022},
url={https://openreview.net/forum?id=nax3ATLrovW}
}

@article{liu2020graphsage,
  title={GraphSAGE-based traffic speed forecasting for segment network with sparse data},
  author={Liu, Jielun and Ong, Ghim Ping and Chen, Xiqun},
  journal={IEEE Transactions on Intelligent Transportation Systems},
  volume={23},
  number={3},
  pages={1755--1766},
  year={2020},
  publisher={IEEE}
}

@inproceedings{gcn,
  author       = {Thomas N. Kipf and
                  Max Welling},
  title        = {Semi-Supervised Classification with Graph Convolutional Networks},
  booktitle    = {5th International Conference on Learning Representations, {ICLR} 2017,
                  Toulon, France, April 24-26, 2017, Conference Track Proceedings},
  publisher    = {OpenReview.net},
  year         = {2017},
  url          = {https://openreview.net/forum?id=SJU4ayYgl}
}

@misc{graphsage,
      title={Inductive Representation Learning on Large Graphs}, 
      author={William L. Hamilton and Rex Ying and Jure Leskovec},
      year={2018},
      eprint={1706.02216},
      archivePrefix={arXiv},
      primaryClass={cs.SI},
      url={https://arxiv.org/abs/1706.02216}, 
}

@article{microsoftgraph,
    author = {Wang, Kuansan and Shen, Zhihong and Huang, Chiyuan and Wu, Chieh-Han and Dong, Yuxiao and Kanakia, Anshul},
    title = {Microsoft Academic Graph: When experts are not enough},
    journal = {Quantitative Science Studies},
    volume = {1},
    number = {1},
    pages = {396-413},
    year = {2020},
    month = {02},
    issn = {2641-3337},
    doi = {10.1162/qss_a_00021},
    url = {https://doi.org/10.1162/qss\_a\_00021},
    eprint = {https://direct.mit.edu/qss/article-pdf/1/1/396/1760880/qss\_a\_00021.pdf},
}

@misc{bai2019deepequilibriummodels,
      title={Deep Equilibrium Models}, 
      author={Shaojie Bai and J. Zico Kolter and Vladlen Koltun},
      year={2019},
      eprint={1909.01377},
      archivePrefix={arXiv},
      primaryClass={cs.LG},
      url={https://arxiv.org/abs/1909.01377}, 
}

@inproceedings{congestionpredictionnet2019,
  title={CongestionNet: Routing congestion prediction using deep graph neural networks},
  author={Kirby, Robert and Godil, Saad and Roy, Rajarshi and Catanzaro, Bryan},
  booktitle={2019 IFIP/IEEE 27th International Conference on Very Large Scale Integration (VLSI-SoC)},
  pages={217--222},
  year={2019},
  organization={IEEE}
}

@inproceedings{generalizable,
  title={Generalizable cross-graph embedding for GNN-based congestion prediction},
  author={Ghose, Amur and Zhang, Vincent and Zhang, Yingxue and Li, Dong and Liu, Wulong and Coates, Mark},
  booktitle={2021 IEEE/ACM International Conference On Computer Aided Design (ICCAD)},
  pages={1--9},
  year={2021},
  organization={IEEE}
}

@inproceedings{highdefcongestion,
  title={High-definition routing congestion prediction for large-scale FPGAs},
  author={Alawieh, Mohamed Baker and Li, Wuxi and Lin, Yibo and Singhal, Love and Iyer, Mahesh A and Pan, David Z},
  booktitle={2020 25th Asia and South Pacific Design Automation Conference (ASP-DAC)},
  pages={26--31},
  year={2020},
  organization={IEEE}
}

@inproceedings{paintingonplacement,
  title={Painting on placement: Forecasting routing congestion using conditional generative adversarial nets},
  author={Yu, Cunxi and Zhang, Zhiru},
  booktitle={Proceedings of the 56th Annual Design Automation Conference 2019},
  pages={1--6},
  year={2019}
}

@inproceedings{congestionaware,
  title={Congestion-aware global routing using deep convolutional generative adversarial networks},
  author={Zhou, Zhonghua and Zhu, Ziran and Chen, Jianli and Ma, Yuzhe and Yu, Bei and Ho, Tsung-Yi and Lemieux, Guy and Ivanov, Andre},
  booktitle={2019 ACM/IEEE 1st Workshop on Machine Learning for CAD (MLCAD)},
  pages={1--6},
  year={2019},
  organization={IEEE}
}

@ARTICLE{Dreamplace,
  author={Lin, Yibo and Jiang, Zixuan and Gu, Jiaqi and Li, Wuxi and Dhar, Shounak and Ren, Haoxing and Khailany, Brucek and Pan, David Z.},
  journal={IEEE Transactions on Computer-Aided Design of Integrated Circuits and Systems}, 
  title={DREAMPlace: Deep Learning Toolkit-Enabled GPU Acceleration for Modern VLSI Placement}, 
  year={2021},
  volume={40},
  number={4},
  pages={748-761},
  doi={10.1109/TCAD.2020.3003843}}

\appendix

\end{document}